\documentclass{article}
\usepackage[preprint]{neurips_2026}

\usepackage[utf8]{inputenc}
\usepackage[T1]{fontenc}
\usepackage{hyperref}
\usepackage{url}
\usepackage{booktabs}
\usepackage{amsmath,amssymb,amsfonts,mathtools}
\usepackage{microtype}
\usepackage{xcolor}
\usepackage{graphicx}
\usepackage{multirow}
\usepackage{siunitx}
\usepackage{array}
\usepackage{caption}
\usepackage{subcaption}
\usepackage{enumitem}
\usepackage[most]{tcolorbox}
\usepackage{pgfplots}
\usepackage{pgfplotstable}
\usepgfplotslibrary{groupplots,colormaps}
\pgfplotsset{compat=1.18}
\usepackage[table]{xcolor}
\usepackage{float}
\definecolor{bestblue}{RGB}{221,235,247}
\definecolor{gaingreen}{RGB}{226,239,218}
\definecolor{lossred}{RGB}{252,228,214}

\newcommand{\bestcell}[1]{\cellcolor{blue!12}\textbf{#1}}
\newcommand{\secondbestcell}[1]{\cellcolor{yellow!18}#1}
\newcommand{\gain}[1]{\cellcolor{gaingreen}\textbf{#1}}
\newcommand{\loss}[1]{\cellcolor{lossred}#1}

\newcommand{\method}{SEB-Cal}

\newcommand{\policy}{Policy-selected}

\title{Beyond Output-Space Calibration: Spectral Evidence Bundling for Selective Reliability Estimation in Time-Series Classification}

\author{
Filippo Cenacchi, Longbing Cao, and Runze Yang\\
Macquarie University, Sydney, Australia\\
\texttt{filippo.cenacchi@mq.edu.au, longbing.cao@mq.edu.au, runze.yang@hdr.mq.edu.au}
}

\begin{document}
\maketitle

\begin{abstract}
Post-hoc calibration for time-series classification usually remaps output scores, yet deployment decisions such as trust, abstention, and review depend on whether a confident prediction is supported by the current temporal signal. This creates three time-series-specific gaps: identical confidence values can hide different temporal support, average calibration can miss false high-confidence errors, and output-space recalibrators provide limited input-linked auditability. We introduce \method{}, a validation-gated fixed-label reliability policy that leaves the backbone and predicted label unchanged while estimating whether that prediction should be trusted. To target these gaps, \method{} augments output-side cues with deterministic whole-sample spectral descriptors: band energy tracks evidence concentration, entropy captures diffusion, peak dominance measures periodic support, and phase stability measures coherent alignment, producing a scalar reliability estimate and band-level diagnostic. The validation gate is a held-out model-selection rule: it enables spectral conditioning only when correctness ranking improves without violating FalseConf@0.9 or AURC tolerances, otherwise reverting to the safer output-space baseline. Across eight heterogeneous UCR/UEA datasets, eight time-series backbone families, and standard output-space recalibrators, unconstrained \method{} improves fixed-label selective-reliability metrics on the matched evaluation subset, increasing Corr-AUROC from \(0.693\) to \(0.779\), while the validation-gated policy further improves Corr-AUROC to \(0.786\) and reduces FalseConf@0.9 to \(0.094\).
\end{abstract}

\section{Introduction}

\begin{figure}[t]
\centering
\includegraphics[width=0.95\textwidth]{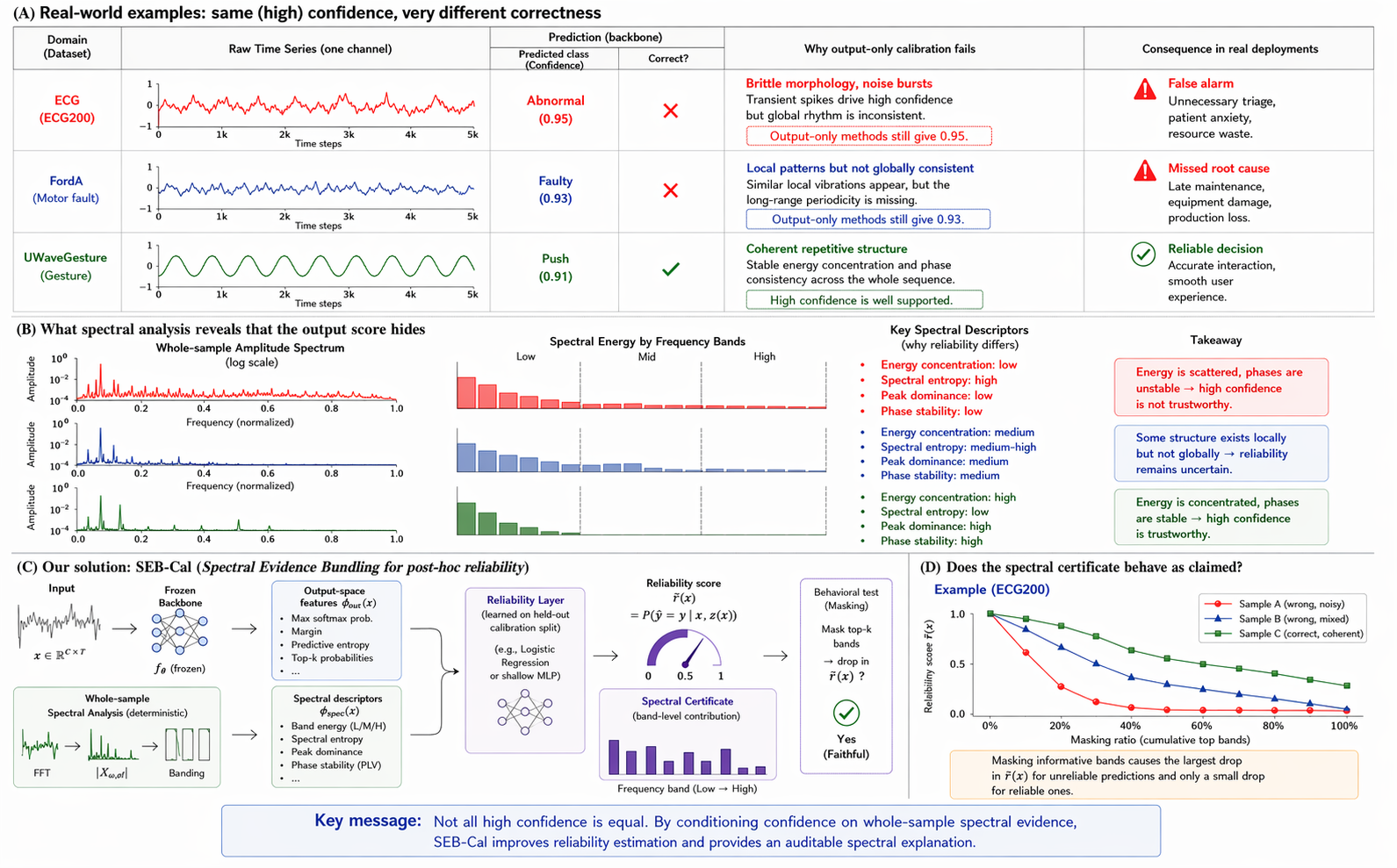}
\caption{
Spectral evidence bundling for post-hoc reliability.
(A) Equally high-confidence predictions can have different signal structures and correctness outcomes.
(B) Spectral analysis reveals energy concentration, entropy, peak dominance, and phase stability missed by output scores.
(C) SEB-Cal conditions confidence on spectral evidence through a post-hoc reliability layer, producing a reliability estimate and band-level diagnostic.
(D) Masking top-ranked spectral bands tests whether reliability follows the diagnostic.
}
\label{fig:spectral_motivation}
\vspace{-1.6em}
\end{figure}

Time-series classification systems are increasingly deployed in workflows where confidence triggers trust, deferral, escalation, filtering, or downstream review. The operational problem is therefore not only whether a classifier is accurate, but whether its confidence tracks correctness for the current sample. This is difficult in time series because two predictions can have identical output-space sharpness while being supported by very different temporal evidence. Most post-hoc calibration methods still operate almost entirely in the output space. Temperature scaling \cite{naeini2015obtaining}, isotonic regression \cite{guo2017calibration}, beta calibration \cite{kull2017beta}, and Dirichlet calibration \cite{kull2019dirichlet} remap logits or probabilities after training and can improve average probability alignment. However, they do not explicitly model whether the current sequence exhibits whole-sample organization: concentrated frequency energy, low spectral diffusion, dominant periodic components, and stable phase alignment. A score of \(0.92\) may arise from coherent long-range organization or from brittle, noisy, locally salient evidence. This matters because time-series reliability is shaped not only by score sharpness, but also by temporal dependence, sequence morphology, and the distribution of evidence across time and frequency. A frequency-based view is therefore attractive because recent time-series models show that periodicity, patch-level structure, global dependencies, and frequency-domain energy compaction are central to temporal representation \cite{wu2023timesnet,nie2023patchtst,yi2023frets,liu2024itransformer}, and because it compactly describes whether signal energy is globally organized across the whole sample set rather than only locally activated. The phenomenon we test is evidence discrepancy: output scores can be high while the input lacks the global temporal organization that should make such confidence reliable. Figure~\ref{fig:spectral_motivation} should be read left-to-right as problem, spectral audit, reliability layer, and masking diagnostic. Band energy locates evidence mass, entropy measures whether it is diffuse or concentrated, peak dominance tests whether a few periodic components carry the signal, and phase stability tests whether components are coherently aligned rather than phase-dispersed. These descriptors are not proposed as universal predictors of correctness; they are probes for a specific failure mode in which score sharpness and temporal support disagree. \method{} therefore acts as a validation-gated reliability policy: spectral evidence is enabled exactly when the held-out data show that this discrepancy signal improves ranking without violating high-confidence error constraints.

We address these gaps through \method{}, a validation-gated fixed-label reliability policy for time-series classification. The backbone is frozen, the predicted label is unchanged, and only the reliability of that prediction is estimated on held-out data. The spectral bundle summarizes each sequence through band-wise energy concentration, spectral entropy, peak dominance, and phase stability computed from a single whole-sample Fourier transform. These descriptors are not used to improve class separation; they test whether the backbone's confidence is supported by globally organized spectral evidence. The contribution of the paper is broader than a new recalibration module. We argue that post-hoc reliability for time-series classification should not be treated purely as an output-space transformation problem. In many temporal settings, whether a high-confidence prediction deserves trust depends partly on how evidence is globally organized in the input itself. \method{} is a concrete instantiation of that thesis: it introduces deterministic whole-sample spectral evidence as a post-hoc reliability signal—distinct from local or learned representations—and exposes the resulting confidence adjustment through a band-level diagnostic stress-tested by masking. The experiments are designed to test exactly that thesis under the conditions that most trouble time-series reliability: heterogeneous temporal dependence, varying sequence morphology, and different distributions of informative evidence across time and frequency. We therefore evaluate across eight datasets from the UCR and UEA archives \cite{dau2019ucr,bagnall2018uea}, eight backbone families spanning recurrent, convolutional, residual, inception-style, and transformer-style predictors \cite{wang2017time,bai2018tcn,ismail2020inceptiontime,vaswani2017attention}, and six standard post-hoc recalibrators beyond the raw confidence score. Unlike a conventional calibration benchmark that asks only whether output scores better match correctness on average, our evaluation asks three complementary questions: whether the method remains competitive on general calibration quality, whether it improves correctness-aware reliability ranking in deployment-relevant settings, and whether its diagnostic has measurable behavioral alignment under intervention. The empirical result is a selective-reliability result with an explicit deployment rule. On the matched evaluation subset, \method{} improves the fixed-label reliability metrics central to selective prediction, including correctness-aware ranking, calibration error, negative log-likelihood, false high-confidence error rate, and retained-risk behavior, with full values reported in Table~\ref{tab:headline_summary}. Accuracy and macro-F1 are reported only as frozen-backbone descriptors because \method{} does not alter the predicted label. At the dataset and backbone levels, the gains are structured rather than uniform: FordA, ECG200, UWaveGestureLibrary, SelfRegulationSCP1, and attention/recurrent backbones provide the clearest positive regimes, while other settings are rejected by the validation gate. This pattern is the intended behavior of the method: spectral evidence is used when it improves reliability under safety constraints, and rejected when output-space confidence remains the safer decision signal. 

Our contributions are fourfold: (i) we identify evidence discrepancy as a reliability failure mode in time-series classification, where output confidence can be high despite weak global temporal support for the selected prediction; (ii) we instantiate this idea through \method{}, a fixed-label evidence-conditioned reliability policy that uses deterministic spectral probes of energy concentration, entropy, peak dominance, and phase stability to decide whether output confidence is temporally supported; (iii) we introduce a deployment-constrained gate that converts regime dependence into an explicit selection rule, using spectral reliability only when held-out validation improves correctness-aware ranking without exceeding FalseConf@0.9 or retained-risk tolerances; and (iv) we provide a matched empirical study across eight datasets, eight backbone families, scalar recalibrators, non-output-space comparators, spectral component ablations, time-frequency variants, time-domain controls, and masking tests.

\vspace{-8pt}
\section{Related Work}
\vspace{-8pt}

Post-hoc calibration converts raw model scores into probabilities that better match empirical correctness. Temperature scaling, isotonic regression, beta calibration, vector scaling, and Dirichlet calibration are widely used because they are modular and often improve probability alignment without retraining the backbone \cite{naeini2015obtaining,guo2017calibration,kull2017beta,kull2019dirichlet}. Benchmark studies show that calibration behavior varies across architectures, datasets, and metrics \cite{tao2023benchmark,lecoz2024many}, suggesting that output scores alone may be incomplete. Sample-dependent methods such as ProCal incorporate geometry or neighborhood structure \cite{xiong2023procal}; \method{} instead conditions reliability on deterministic input-side descriptors of whole-sample temporal organization. The time-series literature increasingly highlights the value of frequency-domain perspectives. Models such as TimesNet, PatchTST, and iTransformer show that strong performance often depends on capturing periodicity, long-range structure, and organized variation that are not easily represented through naive pointwise temporal processing \cite{wu2023timesnet,nie2023patchtst,liu2024itransformer}. Frequency-focused approaches sharpen this intuition. FreTS argues that frequency-domain learning provides a global view and effective energy compaction \cite{yi2023frets}, while FilterNet uses learned frequency filters to exploit the spectrum more selectively \cite{yi2024filternet}. At the same time, recent work warns against simplistic interpretations of Fourier features and emphasizes that their value depends on how they are handled and interpreted \cite{yang2024fbm}. Our use of frequency differs from predictive modeling work: we do not redesign the backbone around the spectrum, nor do we claim that a global Fourier representation is universally sufficient. Instead, we ask whether a compact deterministic summary of whole-sample spectral organization can function as a post-hoc reliability signal. Selective prediction and failure-detection work treats confidence as a decision signal for abstention, triage, and retained-risk control rather than only probability reporting \cite{elyaniv2010selective,geifman2017selective,traub2024augrc,zhu2023openmix}. This motivates our setting: unlike standard post-hoc calibration, which acts only on output scores, \method{} conditions reliability on both output-side calibration cues and whole-sample spectral evidence. The formal fixed-label pipeline is defined in Eq.~\eqref{eq:full_pipeline}; its design rationale is that score sharpness alone may miss whether the signal is globally organized, concentrated, and stable in ways that support trust in the prediction. Explanation adds a further requirement. In high-stakes settings, it is not enough to provide a descriptive visualization disconnected from the confidence signal actually used in decision-making \cite{rudin2019stop,carvalho2019machine}. \method{} therefore explains only the reliability adjustment itself, not the semantic class decision. That narrower target is intentional. Because the diagnostic is derived from the same spectral features that drive the reliability estimate, its relevance can be tested behaviorally through masking: bands ranked as strongly supportive should, when suppressed, induce larger drops in calibrated reliability than lower-ranked bands. The interpretability claim is therefore not real-world causality, but intervention-tested alignment with the reliability adjustment.

\section{Problem Setup}

We define the fixed-label reliability target estimated by \method{}. Let \(x\in\mathbb{R}^{C\times T}\) be a multivariate time series, \(f_{\theta}\) a frozen classifier, \(z(x)\in\mathbb{R}^{K}\) its logits, \(\hat y(x)=\arg\max_k z_k(x)\), and \(p(x)=\mathrm{softmax}(z(x))\). The post-hoc pipeline is
{\setlength{\abovedisplayskip}{2pt}
\setlength{\belowdisplayskip}{2pt}
\setlength{\jot}{1pt}
\tiny
\begin{equation}
\begin{aligned}
z(x)&=f_{\theta}(x), &
\hat y(x)&=\arg\max_k z_k(x), &
p(x)&=\mathrm{softmax}(z(x)),\\
h(x)&=\phi_{\mathrm{out}}(z(x)), &
g(x)&=\phi_{\mathrm{spec}}(x), &
\phi(x)&=[h(x)\Vert g(x)],\\
\bar{\phi}(x)&=\mathrm{Std}(\phi(x)), &
\tilde r(x)&=\psi(\bar{\phi}(x))\approx \mathbb{P}(\hat y(x)=y\mid x,z(x)).
\end{aligned}
\label{eq:full_pipeline}
\end{equation}
}
Here, \(\tilde r(x)\) is the estimated correctness probability, \(\bar{\phi}(x)\) is the standardized concatenated feature vector, and \(\psi\) is the post-hoc reliability model. The backbone and predicted label are fixed; only the reliability layer is learned on held-out calibration data.

\subsection{Reliability Target: Correctness Estimation Rather Than Multiclass Recalibration}

The target of \method{} is deliberately different from the target of multiclass probability recalibration. Standard recalibrators transform a probability vector so that class probabilities better match empirical frequencies. \method{} instead estimates a scalar correctness probability for the already-selected label in Eq.~\eqref{eq:full_pipeline}. Thus, \(\tilde r(x)\) should be interpreted as a fixed-label reliability score, not as a replacement class-probability vector. This distinction determines the evaluation protocol. Ranking metrics such as Corr-AUROC and risk-coverage behavior evaluate whether \(\tilde r(x)\) orders correct predictions above errors. FalseConf@0.9 evaluates whether the method assigns dangerously high reliability to wrong predictions. Binary NLL and binary Brier evaluate the probability quality of the scalar correctness estimate against the binary correctness target \(c_i=\mathbb{I}[\hat y(x_i)=y_i]\). We therefore report probability-quality metrics only on the fixed-label correctness event, while accuracy and macro-F1 remain frozen-backbone descriptors rather than post-hoc method outcomes.

\subsection{Whole-Sample Spectral Evidence Bundling}

The spectral branch is designed to test whether correctness depends on global organization that is not visible in the output scores. The bundle is intentionally deterministic and low-capacity: it does not learn a new representation of the time series, but summarizes four complementary properties of whole-sample temporal evidence. Band energy measures where spectral mass is concentrated; entropy measures whether the evidence is diffuse or organized; peak dominance measures whether a few dominant periodic components explain the signal; and phase stability measures whether spectral components are coherently aligned across channels and bands. This is why the spectral branch is gated rather than universal: STFT, time-domain, and masking controls separate global spectral support from local structure, generic input statistics, and perturbation sensitivity. For each sample we compute a single whole-sample real Fourier transform per channel. Let
\begin{equation}
\tiny
X_{c,\omega} = \mathcal{F}(x_{c,:})_{\omega},
\label{eq:fft}
\end{equation}
where \(X_{c,\omega}\) is the complex Fourier coefficient for channel \(c\) and frequency index \(\omega\). From this full-spectrum representation we derive a compact evidence bundle. The first component is band-wise energy concentration. In all experiments we use
\(B=8\) disjoint frequency bands. For a length-\(T\) sequence, the real FFT returns
\(F=\lfloor T/2\rfloor+1\) non-negative frequency coefficients. We exclude the DC
coefficient from the band partition and define the positive-frequency index set
\(\Omega^{+}=\{1,\dots,F-1\}\). The ordered positive-frequency indices are split into
\(B\) contiguous, approximately equal-width index bands,
\begin{equation}
\tiny
\mathcal{B}_b =
\left\{
\omega \in \Omega^{+} :
1+\left\lfloor \frac{(b-1)|\Omega^{+}|}{B}\right\rfloor
\leq \omega \leq
\left\lfloor \frac{b|\Omega^{+}|}{B}\right\rfloor
\right\},
\quad b=1,\dots,B.
\label{eq:frequency_bands}
\end{equation}
If \(|\Omega^{+}|<B\), empty bands are omitted and the remaining bands are indexed in
ascending frequency order. We use frequency-index partitioning rather than dataset-specific
physical frequency cutoffs because UCR/UEA datasets do not consistently provide sampling
rates. We then compute
\begin{equation}
\tiny
e_b(x)=\log\!\left(1+\sum_{c=1}^{C}\sum_{\omega\in\mathcal{B}_b}\lvert X_{c,\omega}\rvert^2\right),
\label{eq:bandenergy}
\end{equation}
where \(e_b(x)\) is the log energy in band \(b\). These band energies summarize where spectral mass concentrates. The second component is spectral entropy. We first define the normalized spectral weights
\begin{equation}
\tiny
q_{\omega}(x)=\frac{\sum_{c=1}^{C}\lvert X_{c,\omega}\rvert^2}{\sum_{\omega'}\sum_{c=1}^{C}\lvert X_{c,\omega'}\rvert^2+\epsilon},
\label{eq:qomega}
\end{equation}
where \(q_{\omega}(x)\) is the normalized energy at frequency \(\omega\), and then compute
\begin{equation}
\tiny
H_{\mathrm{spec}}(x)=
-\frac{1}{\log |\Omega^{+}|}\sum_{\omega\in\Omega^{+}} q_{\omega}(x)\log\big(q_{\omega}(x)+\epsilon\big),
\label{eq:specentropy}
\end{equation}
where \(|\Omega^{+}|\) is the number of retained positive-frequency coefficients and
\(H_{\mathrm{spec}}(x)\) is the normalized spectral entropy. The third component is peak dominance. We first aggregate complex amplitudes across channels
at each retained positive frequency by the root-sum-square magnitude
\begin{equation}
\tiny
a_{\omega}(x)=
\left(\sum_{c=1}^{C}\lvert X_{c,\omega}\rvert^2\right)^{1/2},
\qquad \omega \in \Omega^{+}.
\label{eq:channel_aggregated_amplitude}
\end{equation}
The values \(\{a_{\omega}(x):\omega\in\Omega^{+}\}\) are sorted as
\(a_{(1)} \ge a_{(2)} \ge \dots\), with unavailable order statistics set to zero. We compute
\begin{equation}
\tiny
d_1(x)=\frac{a_{(1)}}{\sum_j a_{(j)}+\epsilon}, \qquad
d_3(x)=\frac{\sum_{j=1}^{3}a_{(j)}}{\sum_j a_{(j)}+\epsilon}, \qquad
d_5(x)=\frac{\sum_{j=1}^{5}a_{(j)}}{\sum_j a_{(j)}+\epsilon},
\label{eq:peakdom}
\end{equation}
where \(d_1(x)\), \(d_3(x)\), and \(d_5(x)\) measure how much total spectral mass is concentrated in the largest one, three, and five peaks. The fourth component is phase stability. For each band, we compute
\begin{equation}
\tiny
\kappa_b(x)=
\left\lvert
\frac{1}{C\lvert\mathcal{B}_b\rvert}
\sum_{c=1}^{C}\sum_{\omega\in\mathcal{B}_b}\exp(i\,\angle X_{c,\omega})
\right\rvert,
\label{eq:phasestability}
\end{equation}
where \(\kappa_b(x)\) is the concentration of unit phasors in band \(b\), and therefore summarizes within-band phase coherence. We concatenate these descriptors into the spectral evidence bundle
\begin{equation}
\tiny
g(x)=\big[e_1(x),\dots,e_B(x),\, H_{\mathrm{spec}}(x),\, d_1(x),d_3(x),d_5(x),\, \kappa_1(x),\dots,\kappa_B(x)\big],
\label{eq:bundle}
\end{equation}
where \(g(x)\) is the deterministic feature vector supplied to the reliability layer. The reliability layer is deliberately low-capacity (logistic regression or a small MLP) so that success cannot be attributed to a powerful auxiliary learner. This makes the test conservative: any gain must come from exposing score--evidence discrepancy through the spectral bundle rather than from relearning the classifier, consistent with evidence that stronger time-series performance does not necessarily require increasingly complex auxiliary modules~\cite{zeng2023transformers}. Thus, the default spectral bundle has \(2B+4=20\) features when all eight bands are
non-empty: \(B\) band-energy values, one normalized spectral entropy value, three
peak-dominance values, and \(B\) phase-stability values.

\subsection{Output Features and Reliability Model}

Because the spectral bundle is intended to complement rather than replace standard calibration cues, the reliability model combines it with conventional output-side features derived from the backbone logits. The output feature vector \(h(x)\) consists of maximum softmax probability, logit margin, and predictive entropy as defined in Eq.~\eqref{eq:hx}.
\begin{align}
\tiny
h(x) = \big[
\underbrace{\max_k p_k(x)}_{\text{MSP } s(x)},\;
\underbrace{z_{(1)}(x)-z_{(2)}(x)}_{\text{margin } m(x)},\;
\underbrace{-\sum_{k=1}^{K} p_k(x)\log(p_k(x)+\epsilon)}_{\text{entropy } H_{\mathrm{pred}}(x)}
\big].
\label{eq:hx}
\end{align}
where \(z_{(1)}(x)\) and \(z_{(2)}(x)\) are the largest and second-largest logits. These quantities form the output feature vector \(h(x)\). The final reliability model operates on the concatenated feature vector
\begin{equation}
\tiny
\phi(x)=\big[h(x)\,\Vert\,g(x)\big],
\label{eq:featureconcat}
\end{equation}
where \(\Vert\) denotes vector concatenation. In the default implementation, \(\phi(x)\) is standardized and passed to a shallow logistic reliability model,
\begin{equation}
\tiny
\tilde{r}(x)=\sigma\big(w^\top \bar{\phi}(x)+b\big),
\label{eq:reliabilitymodel}
\end{equation}
where \(\bar{\phi}(x)\) is the standardized feature vector, \(w\) and \(b\) are learned parameters, and \(\sigma(\cdot)\) is the logistic sigmoid. Training uses binary correctness targets
\begin{equation}
\tiny
c_i=\mathbb{I}\big[\hat{y}(x_i)=y_i\big],
\label{eq:correctness}
\end{equation}
where \(c_i=1\) if the backbone prediction is correct and \(0\) otherwise. The loss is the binary cross-entropy
\begin{equation}
\tiny
\mathcal{L}_{\mathrm{cal}}=
-\frac{1}{n}\sum_{i=1}^{n}
\Big(c_i\log\tilde{r}(x_i)+(1-c_i)\log(1-\tilde{r}(x_i))\Big),
\label{eq:loss}
\end{equation}
where \(n\) is the number of calibration samples. Eqs.~\eqref{eq:correctness}--\eqref{eq:loss} define the learning objective for the post-hoc reliability layer.  The output \(\tilde r(x)\) is not a class probability vector and is never used to change \(\hat y(x)\). It is a scalar estimate of whether the already-selected label is correct. Consequently, any method-specific difference in the paper's comparative results must come from reliability scores, not from changed predictions. Accuracy and macro-F1 are therefore frozen-backbone descriptors rather than post-hoc method outcomes.

\paragraph{Validation-gated reliability policy.}
The deployed object is a held-out selection operator, not an unconstrained spectral score. A spectral reliability model is selected only when it satisfies
{\tiny
\begin{align}
\Delta \mathrm{CorrAUROC} &> \delta_{\mathrm{rank}}, \\
\Delta \mathrm{FalseConf@0.9} &\leq \tau_{\mathrm{fc}}, \\
\Delta \mathrm{AURC} &\leq \tau_{\mathrm{aurc}},
\end{align}
}
relative to Raw and the strongest scalar recalibrator on the same gate-validation split. Unless otherwise stated, thresholds are fixed before final test evaluation: \(\delta_{\mathrm{rank}}>0\), \(\tau_{\mathrm{fc}}=0.05\) for low-to-moderate baseline FalseConf regimes, and \(\tau_{\mathrm{fc}}=0.08\) when Raw FalseConf@0.9 exceeds \(0.15\); increases above \(0.10\) absolute are deployment-incompatible regardless of ranking gain. If the constraints fail, the selected policy reverts to Raw confidence or the best simpler output-space recalibrator. Thus, \method{} contributes both a spectral reliability score and a pre-specified policy for when that score may safely replace simpler recalibration. Dataset-level gate outcomes and the policy-selected aggregate comparison are reported in Appendix~\ref{app:deployment_selection}.

\subsection{Spectral Diagnostic and Faithfulness Test}

To make the reliability adjustment inspectable, the method produces a per-sample spectral diagnostic and tests whether that diagnostic is behaviorally aligned with the reliability score under masking. The diagnostic is not presented as a semantic explanation of the class decision, nor as a causal explanation of the data-generating process. Its narrower purpose is to expose which spectral bands contributed to the post-hoc reliability adjustment and to test whether suppressing those bands changes the estimated correctness probability in the expected direction. Let \(w_e,w_\kappa,w_H,w_{d_q}\) denote the learned coefficients on standardized spectral features. We assign only band-indexed terms to the band diagnostic:
\begin{equation}
\tiny
A_b(x)=w_{e_b}\bar e_b(x)+w_{\kappa_b}\bar\kappa_b(x),\qquad
A_{\mathrm{glob}}(x)=w_H\bar H_{\mathrm{spec}}(x)+\sum_{q\in\{1,3,5\}}w_{d_q}\bar d_q(x).
\label{eq:band_certificate}
\end{equation}
Only \(A_b(x)\) is ranked for masking; the global entropy and peak-dominance contribution \(A_{\mathrm{glob}}(x)\) affects \(\tilde r(x)\) but is not assigned to individual bands. For input-space faithfulness, bands are suppressed and the reliability drop is
\begin{equation}
\tiny
\Delta_{\mathcal{S}}(x)=\tilde{r}(x)-\tilde{r}(x^{(-\mathcal{S})}),
\label{eq:drop}
\end{equation}
where \(x^{(-\mathcal{S})}\) is obtained after zeroing Fourier coefficients in \(\mathcal{S}\). If faithful, larger \(A_b(x)\) should correspond to larger drops in Eq.~\eqref{eq:drop}. Because this end-to-end test can include frozen-backbone sensitivity, Appendix~\ref{app:feature_mask_faithfulness} reports a feature-space masking control with \(h(x)\) fixed.

\section{Experimental Protocol}

To test regime dependence rather than benchmark-specific behavior, we evaluate eight heterogeneous UCR/UEA datasets \cite{dau2019ucr,bagnall2018uea}: \textsc{ECG200}, \textsc{FordA}, \textsc{Wafer}, \textsc{ElectricDevices}, \textsc{UWaveGestureLibrary}, \textsc{BasicMotions}, \textsc{SelfRegulationSCP1}, and \textsc{AtrialFibrillation}. They span univariate/multivariate inputs, binary/multiclass labels, short/long sequences, and physiology, industrial sensing, and human-motion domains. All splits, seeds, trained-output files, table-generation scripts, paper-value CSVs, and README-level reproduction figures are fixed before test evaluation and documented in the anonymized artifact described in Appendix~\ref{app:artifact_addendum}. The artifact separates fast verification from full reproduction: reviewers can inspect paper-value CSVs, README figures, and table outputs directly before rerunning the fixed-split benchmark from code. We evaluate \method{} across eight backbone families and canonical post-hoc baselines to test whether spectral evidence adds reliability information beyond architecture effects or score remapping. Backbones are MLP, LSTM, GRU, TCN, FCN, 1D ResNet, lightweight Inception-style classifier, and Transformer encoder \cite{wang2017time,bai2018tcn,ismail2020inceptiontime,vaswani2017attention}. Baselines are raw confidence, temperature scaling \cite{guo2017calibration}, Platt-style logistic calibration, isotonic regression, beta calibration \cite{kull2017beta}, vector scaling, and Dirichlet calibration \cite{kull2019dirichlet}. They test whether deterministic input-side evidence contributes reliability information not recoverable by score transformation alone. Proximity, time-domain, STFT, masking, and overhead controls are reported in Appendix~\ref{app:extended_robustness}. 

Because \method{} is fixed-label post-hoc, we separate frozen predictive quality from reliability quality. Accuracy and macro-F1 are computed once from backbone predictions and reported only as task context. All comparative metrics use the scalar reliability score assigned to the fixed prediction. We report: (i) fixed-label probability quality using ECE, binary NLL, and binary Brier against \(c_i=\mathbb{I}[\hat y(x_i)=y_i]\); (ii) decision-time reliability using Corr-AUROC, FalseConf@0.9, and AURC, where Corr-AUROC ranks correct predictions above errors, FalseConf@0.9 is the fraction of errors assigned reliability above \(0.9\), and AURC measures retained risk under selective coverage; and (iii) diagnostic faithfulness using Spearman correlation between band-level diagnostic scores and reliability drops after masking. Random-band, equal-energy, and leave-one-band-out controls are reported in the appendix to test whether the diagnostic captures reliability-relevant spectral structure rather than generic perturbation sensitivity or energy magnitude. 

\section{Results and Discussion}

\subsection{Main Results}

We organize the evaluation along the same dimensions introduced in the experimental design: aggregate reliability quality, dataset-level heterogeneity, backbone-family variation, diagnostic scope ablations, and spectral diagnostic behavior. Results are reported on the matched evaluation subset of \(191\) dataset--model--seed configurations for which all standard post-hoc methods are available, yielding directly comparable reliability evaluations. Table~\ref{tab:headline_summary} provides the strictest aggregate comparison across heterogeneous datasets, architectures, and seeds. Because all post-hoc methods keep the predicted label fixed, it compares only reliability metrics; frozen backbone accuracy and macro-F1 are shown once in the caption for context, not as method-specific outcomes. Unconstrained \method{} improves the core fixed-label reliability metrics over Raw and scalar recalibrators, including Corr-AUROC, ECE, NLL, FalseConf@0.9, AURC, and the spectral diagnostic faithfulness score. The validation-gated policy is the deployed object: it further improves Corr-AUROC, ECE, NLL, FalseConf@0.9, and AURC by selecting spectral reliability only when held-out validation passes the ranking and safety constraints. Thus, Table~\ref{tab:headline_summary} separates two claims: the spectral bundle adds reliability information beyond output-score remapping, and the gate converts that regime-dependent signal into a safer operating policy. \textbf{Uncertainty.} All aggregate comparisons are computed over matched dataset--backbone--seed configurations, with paired bootstrap uncertainty reported in Appendix~\ref{app:bootstrap}. Bootstrap intervals support stability only within the reported evidence base, since dataset and backbone decompositions remain heterogeneous. Appendix~\ref{app:extended_robustness} separates spectral organization from generic input conditioning: \(g(x)\) has low dependence on \(h(x)\) (median absolute Spearman \(<0.25\)), exceeds a time-domain summary control, and is stress-tested against Proximity and STFT variants. Together, these controls rule out the weakest interpretation of the result: that \method{} merely benefits from adding any input-derived side information to a shallow correctness model. The evidence instead supports a narrower and stronger claim: frequency-distributed temporal organization contributes reliability information specifically when confidence errors reflect a mismatch between output sharpness and global signal support.

\begin{table}[H]
\centering
\vspace{-0.9em}
\tiny
\caption{Aggregate fixed-label reliability comparison on the matched evaluation subset. Backbone predictions are fixed for all post-hoc methods, so accuracy and macro-F1 are not method-specific; frozen predictive context is accuracy \(=0.701\), macro-F1 \(=0.636\). Higher is better for Corr-AUROC and Faithfulness; lower is better for ECE, Brier, NLL, FalseConf@0.9, and AURC.}
\label{tab:headline_summary}
\begin{tabular}{lccccccc}
\toprule
Method & ECE & Brier & NLL & Corr-AUROC & FalseConf@0.9 & AURC & Faith. \\
\midrule
Raw         
& 0.097 
& 0.377 
& \secondbestcell{0.716} 
& \secondbestcell{0.693} 
& 0.128 
& 0.219 
& -- \\

Temperature 
& 0.080 
& \bestcell{0.371} 
& 0.717 
& 0.689 
& 0.162 
& 0.219 
& -- \\

Platt       
& \secondbestcell{0.078} 
& 0.384 
& 0.745 
& 0.671 
& 0.228 
& \secondbestcell{0.218} 
& -- \\

Dirichlet   
& 0.153 
& 0.422 
& 0.983 
& 0.673 
& 0.153 
& 0.226 
& -- \\

Vector      
& 0.155 
& 0.420 
& 0.975 
& 0.672 
& 0.149 
& 0.227 
& -- \\

Isotonic    
& 0.144 
& 0.421 
& 1.977 
& 0.644 
& 0.292 
& 0.227 
& -- \\

Unconstrained \method{}     
& \secondbestcell{0.077} 
& 0.372 
& \secondbestcell{0.710} 
& \secondbestcell{0.779} 
& \secondbestcell{0.118} 
& \secondbestcell{0.214} 
& \bestcell{0.059} \\

\policy{} \method{}     
& \bestcell{0.075} 
& \bestcell{0.370} 
& \bestcell{0.708} 
& \bestcell{0.786} 
& \bestcell{0.094} 
& \bestcell{0.209} 
& -- \\
\bottomrule
\end{tabular}
\vspace{-1.3em}
\end{table}

\subsection{Cross-Dataset Heterogeneity \& Backbone-Level Evidence}

Table~\ref{tab:dataset_table} separates ranking gains from high-confidence error behavior. FordA, ECG200, UWaveGestureLibrary, and SelfRegulationSCP1 show clear Corr-AUROC gains, while BasicMotions and AtrialFibrillation remain unfavorable. Wafer and ElectricDevices illustrate the main operational tradeoff: spectral evidence can improve ranking even when FalseConf@0.9 must be checked against the deployment tolerance. The full validation-gate decision table is moved to Appendix~\ref{app:deployment_selection}. Across backbones, the same ranking asymmetry appears in Table~\ref{tab:model_table}: Transformer and LSTM are the clearest positive Corr-AUROC cases, whereas convolutional families and the MLP group are weaker. This supports the central claim that regime dependence is the empirical signature of the phenomenon: spectral reliability helps when confidence failures reflect mismatch between output sharpness and global temporal support.

\begin{table}[t]
\centering
\tiny
\caption{Dataset-level decomposition for unconstrained \method{}. Positive Corr-AUROC gains are in green. FalseConf@0.9 is shown diagnostically; deployment safety is handled by the validation-gated policy in Table~\ref{tab:headline_summary} and Appendix~\ref{app:deployment_selection}.}
\label{tab:dataset_table}
\begin{tabular}{lcccccccc}
\toprule
Dataset & Raw Corr & SEB Corr & $\Delta$ Corr & Raw FalseConf & SEB FalseConf & Raw ECE & SEB ECE & Faith. \\
\midrule
FordA & 0.730 & 0.859 & \gain{+0.129} & 0.101 & 0.124 & 0.024 & 0.034 & 0.113 \\
ECG200 & 0.685 & 0.750 & \gain{+0.065} & 0.105 & 0.129 & 0.112 & 0.178 & 0.103 \\
BasicMotions & 0.597 & 0.586 & \loss{-0.011} & 0.037 & 0.083 & 0.081 & 0.079 & 0.027 \\
ElectricDevices & 0.713 & 0.723 & \gain{+0.010} & 0.251 & 0.268 & 0.139 & 0.132 & \bestcell{0.427} \\
AtrialFibrillation & 0.499 & 0.456 & \loss{-0.043} & 0.104 & 0.174 & 0.160 & 0.437 & 0.201 \\
Wafer & 0.826 & 0.885 & \gain{+0.059} & 0.455 & 0.519 & 0.015 & 0.012 & 0.019 \\
UWaveGestureLibrary & 0.685 & 0.754  & \gain{+0.069} & 0.084 & 0.056 & 0.136 & 0.186 & 0.073 \\
SelfRegulationSCP1 & 0.567 & 0.671 & \gain{+0.104} & 0.249 & 0.241 & 0.110 & 0.192 & \loss{-0.505} \\
\bottomrule
\end{tabular}
\vspace{-1.6em}
\end{table}

\begin{table}[t]
\centering
\tiny
\caption{Backbone-level decomposition for unconstrained \method{}. Positive Corr-AUROC gains are in green. High-confidence error safety is evaluated at the aggregate and policy levels in Table~\ref{tab:headline_summary} and Appendix~\ref{app:deployment_selection}.}
\label{tab:model_table}
\begin{tabular}{lccccc}
\toprule
Backbone & Raw Corr & SEB Corr & $\Delta$ Corr & SEB ECE & Faith. \\
\midrule
Transformer & 0.583 & 0.699 & \gain{+0.116} & 0.136 & 0.084 \\
LSTM & 0.659 & 0.733 & \gain{+0.074} & 0.168 & \loss{-0.002} \\
GRU & 0.676 & 0.667 & \loss{-0.009} & 0.206 & 0.039 \\
ResNet1D & 0.702 & 0.672 & \loss{-0.030} & 0.141 & 0.055 \\
TCN & 0.704 & 0.670 & \loss{-0.034} & 0.142 & 0.043 \\
FCN & 0.707 & 0.671 & \loss{-0.036} & 0.153 & 0.105 \\
MLP & 0.762 & 0.672 & \loss{-0.090} & 0.179 & 0.075 \\
InceptionLite & 0.748 & 0.652 & \loss{-0.096} & 0.130 & 0.090 \\
\bottomrule
\end{tabular}
\vspace{-2.3em}
\end{table}

\begin{tcolorbox}[colback=blue!3,colframe=blue!35,title=Validation-gated reliability policy]
\scriptsize
The deployed object is a selected reliability policy, not an unconstrained spectral score. Held-out validation first tests whether spectral evidence improves correctness-aware ranking; it then checks whether high-confidence error risk and retained risk remain within tolerance. When the gate passes, \method{} is selected because input-side temporal organization adds reliability information beyond output scores. When the gate fails, the system reverts to Raw confidence or the safer scalar recalibrator. This turns heterogeneous behavior across datasets into an explicit reliability-selection mechanism.
\end{tcolorbox}

\subsection{Frequency-Bundle Ablation}

Table~\ref{tab:frequency_bundle_ablation} directly ablates the proposed spectral evidence
bundle \(g(x)\). This table is a mechanism test for ranking signal, not a deployment-safety table:
high-confidence error safety is evaluated by the validation gate in Table~\ref{tab:headline_summary}
and Appendix~\ref{app:deployment_selection}. Removing the entire bundle reduces the method to
the output-side reliability model \(h(x)\), while removing individual descriptor families tests which
part of the bundle drives the ranking gain. The full bundle outperforms the output-only model,
showing that \(g(x)\) contributes information beyond logits, margin, and predictive entropy. Among
individual components, energy concentration is the strongest single descriptor, while the leave-one-out
rows show that removing energy causes the clearest drop from the full bundle. Entropy, peak dominance,
and phase stability contribute less uniformly, but preserve a decomposable spectral diagnostic and
allow the reliability adjustment to be inspected as structured evidence rather than as a single opaque
scalar. This ablation addresses the role of \(g(x)\): removing the whole bundle reduces Corr-AUROC
from \(0.786\) to \(0.703\), while the leave-one-out rows identify energy concentration as the dominant
component in this slice. Additional visual summaries of the aggregate trade-off, dataset--backbone
regimes, and diagnostic faithfulness are provided in Appendix~\ref{app:visual_summary}.

\begin{table}[t]
\centering
\tiny
\caption{\textbf{Frequency-bundle ablation.}
Mechanism ablation for ranking signal; FalseConf@0.9 and AURC safety are evaluated by the validation-gated policy in Table~\ref{tab:headline_summary} and Appendix~\ref{app:deployment_selection}. The first block removes \(g(x)\) or adds one component family; the second removes one family from the full bundle. Higher Corr-AUROC is better. Positive regime: FordA + ECG200.}
\label{tab:frequency_bundle_ablation}
\begin{tabular}{lc}
\toprule
Variant & Corr-AUROC \\
\midrule
Remove all \(g(x)\): output-side only \(h(x)\) & 0.703 \\
\(h(x)+\) energy only & 0.778 \\
\(h(x)+\) entropy only & 0.733 \\
\(h(x)+\) peak only & 0.756 \\
\(h(x)+\) phase only & 0.734 \\
Full \method{} bundle \(h(x)+g(x)\) & 0.786 \\
\midrule
Full bundle without energy & 0.752 \\
Full bundle without entropy & 0.786 \\
Full bundle without peak & 0.784 \\
Full bundle without phase & \bestcell{0.787} \\
\bottomrule
\end{tabular}
\vspace{-1.9em}
\end{table}

\subsection{Spectral Diagnostic Behavior}

Finally, we test whether the spectral diagnostic is behaviorally aligned with the reliability adjustment rather than merely descriptive. The diagnostic explains the post-hoc reliability score, not the class decision or real-world causality. If specific spectral bands drive the reliability layer, masking higher-ranked bands should change \(\tilde r(x)\) more than masking lower-ranked bands. We therefore rank bands by diagnostic score and progressively suppress them before recalculating \(\tilde r(x)\); Appendix~\ref{app:feature_mask_faithfulness} reports a feature-space masking control that holds backbone output features fixed. On the matched evaluation subset, the mean rank--drop association is positive but modest, so the diagnostic is secondary; the primary acceptance case rests on fixed-label reliability ranking and validation-gated deployment. Dataset-level variation is informative: most datasets show positive alignment, whereas SelfRegulationSCP1 marks a clear boundary case. When diagnostic validity is required, the deployment rule rejects regimes that fail random-band or equal-energy controls. The diagnostic is therefore validation-gated, not an unconditional interpretability guarantee. The time-domain summary control in Appendix~\ref{app:extended_robustness} further shows that the gain is not explained by adding generic input statistics alone.

\section{Operating Contract and Failure Modes}

The operating contract of \method{} follows from its fixed-label design: it improves the reliability score assigned to an existing backbone prediction but never alters the predicted class. Thus, \(\tilde r(x)\) is evaluated as a scalar correctness probability, with NLL and Brier interpreted against \(c_i=\mathbb{I}[\hat y(x_i)=y_i]\). This target suits selective prediction, triage, review prioritization, and abstention, where the key question is whether the prediction should be trusted. The main failure mode occurs when global spectral organization is uninformative about correctness, especially in tasks dominated by transients, changepoints, padding artifacts, or non-spectral nuisance factors. Ranking gains can also coexist with worse FalseConf@0.9, so they are insufficient for deployment. Finally, the spectral diagnostic explains the post-hoc reliability adjustment, not the semantic class decision or causal structure of the signal, and frequency masking may produce non-natural perturbations. \method{} is therefore enabled only when held-out validation shows improved ranking over Raw and scalar recalibrators, no unacceptable FalseConf@0.9 or AURC increase, and, when the diagnostic is exposed, faithfulness above random-band and equal-energy controls; otherwise it reverts to Raw or best simpler recalibrator. Under this contract, unfavorable regimes are expected rejection cases rather than hidden failures.

\section{Conclusion}

This paper introduced \method{}, a validation-gated fixed-label reliability policy for time-series classification. The method leaves the backbone and predicted label unchanged, augments output-side cues with deterministic spectral evidence, and estimates whether the selected prediction should be trusted. The central finding is that time-series reliability can exhibit evidence discrepancy: output confidence can be high even when global temporal organization provides weak support for correctness. Across heterogeneous datasets and backbones, \method{} exploits this discrepancy signal, improves correctness-aware ranking, and uses a validation gate to decide when spectral reliability should replace simpler output-space recalibration. The result reframes time-series calibration as evidence-constrained reliability selection rather than unconditional score remapping.

\bibliographystyle{unsrt}
\bibliography{references}

\appendix

\section{Appendix: Abbreviations}
\label{app:abbreviations}

\begin{table}[h]
\centering
\scriptsize
\caption{Abbreviations used in the paper.}
\label{tab:abbreviations}
\begin{tabular}{ll}
\toprule
Abbreviation & Definition \\
\midrule
\method{} / SEB-Cal & Spectral Evidence Bundling Calibration \\
SEB & Spectral Evidence Bundling \\
FFT & Fast Fourier Transform \\
STFT & Short-Time Fourier Transform \\
STFT-\method{} & Short-Time Fourier Transform variant of SEB-Cal \\
UCR & University of California Riverside time-series archive \\
UEA & University of East Anglia time-series archive \\
UCR/UEA & UCR and UEA time-series classification archives \\
ECE & Expected Calibration Error \\
NLL & Negative Log-Likelihood \\
AUROC & Area Under the Receiver Operating Characteristic Curve \\
Corr-AUROC & Correctness-aware Area Under the Receiver Operating Characteristic Curve \\
AURC & Area Under the Risk-Coverage Curve \\
FalseConf@0.9 & False high-confidence error rate at reliability threshold 0.9 \\
MSP & Maximum Softmax Probability \\
F1 & Harmonic mean of precision and recall \\
MLP & Multilayer Perceptron \\
LSTM & Long Short-Term Memory network \\
GRU & Gated Recurrent Unit \\
TCN & Temporal Convolutional Network \\
FCN & Fully Convolutional Network \\
ResNet1D & One-dimensional Residual Network \\
CI & Confidence Interval \\
IRB & Institutional Review Board \\
LLM & Large Language Model \\
ECG200 & Electrocardiogram 200 dataset \\
\bottomrule
\end{tabular}
\end{table}

\section{Appendix: Full Per-Dataset Method Comparison}

Table~\ref{tab:appendix_corr} reports the full per-dataset Corr-AUROC comparison on the completed common test subset. Table~\ref{tab:appendix_falseconf} reports the corresponding gate-validation FalseConf@0.9 diagnostics used for policy selection; these values are intentionally separated from the final test-set FalseConf@0.9 values in Table~\ref{tab:dataset_table}.

\begin{table*}[t]
\centering
\scriptsize
\caption{Full per-dataset Corr-AUROC comparison across methods. Highest value in each row is highlighted.}
\label{tab:appendix_corr}
\begin{tabular}{lccccccc}
\toprule
Dataset & Raw & Temp. & Platt & Dirichlet & Vector & Isotonic & SEB-Cal \\
\midrule
AtrialFibrillation & 0.499 & \bestcell{0.501} & 0.489 & 0.455 & 0.440 & 0.498 & 0.456 \\
BasicMotions & \bestcell{0.597} & 0.584 & 0.579 & 0.574 & 0.565 & 0.560 & 0.586 \\
ECG200 & 0.685 & 0.685 & 0.667 & 0.636 & 0.661 & 0.582 & \bestcell{0.750} \\
ElectricDevices & 0.713 & 0.724 & 0.725 & \bestcell{0.727} & 0.721 & 0.722 & 0.723 \\
FordA & 0.730 & 0.730 & 0.730 & 0.735 & 0.730 & 0.726 & \bestcell{0.859} \\
SelfRegulationSCP1 & 0.567 & 0.671 & 0.665 & \bestcell{0.685} & 0.666 & 0.615 & 0.671 \\
UWaveGestureLibrary & 0.685 & 0.739 & 0.752 & 0.727 & 0.733 & 0.700 & \bestcell{0.754} \\
Wafer & 0.826 & 0.865 & 0.766 & 0.851 & 0.865 & 0.754 & \bestcell{0.885} \\
\bottomrule
\end{tabular}
\end{table*}

\begin{table*}[t]
\centering
\scriptsize
\caption{Gate-validation FalseConf@0.9 comparison used only for policy-selection diagnostics. These values are computed on the held-out gate-validation split and must not be compared algebraically with the test-set values in Table~\ref{tab:dataset_table}.}
\label{tab:appendix_falseconf}
\begin{tabular}{lccccccc}
\toprule
Dataset & Raw & Temp. & Platt & Dirichlet & Vector & Isotonic & SEB-Cal \\
\midrule
ElectricDevices & 0.251 & 0.213 & 0.274 & \bestcell{0.203} & 0.210 & 0.239 & 0.268 \\
BasicMotions & 0.037 & 0.062 & 0.042 & 0.150 & 0.083 & 0.179 & \bestcell{0.016} \\
ECG200 & 0.329 & 0.103 & 0.125 & 0.164 & 0.152 & 0.416 & \bestcell{0.075} \\
FordA & 0.101 & 0.076 & 0.132 & 0.061 & 0.065 & 0.124 & \bestcell{0.027} \\
SelfRegulationSCP1 & 0.188 & 0.179 & 0.441 & 0.108 & 0.124 & 0.366 & \bestcell{0.069} \\
UWaveGestureLibrary & 0.019 & 0.091 & 0.208 & 0.277 & 0.005 & 0.237 & \bestcell{0.004} \\
Wafer & 0.355 & 0.556 & 0.967 & 0.373 & 0.519 & 0.478 & \bestcell{0.325} \\
\bottomrule
\end{tabular}
\end{table*}

\section{Appendix: Empirical Visual Summary}
\label{app:visual_summary}

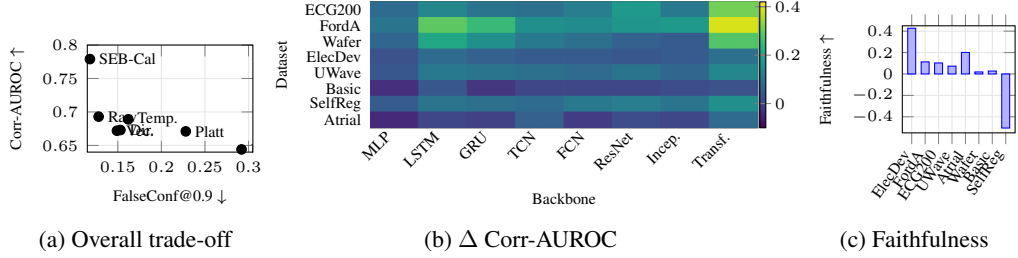
\begin{figure*}[t]
\centering

\begin{subfigure}[t]{0.27\textwidth}
\centering
\begin{tikzpicture}
\begin{axis}[
width=\linewidth,
height=3.0cm,
xlabel={FalseConf@0.9 $\downarrow$},
ylabel={Corr-AUROC $\uparrow$},
xlabel style={font=\tiny},
ylabel style={font=\tiny},
xmin=0.115, xmax=0.305,
ymin=0.64, ymax=0.80,
grid=both,
major grid style={gray!20},
minor grid style={gray!10},
tick label style={font=\tiny},
label style={font=\tiny},
]
\addplot[only marks, mark=*, mark size=1.8pt] coordinates {
(0.128,0.693)
(0.162,0.689)
(0.228,0.671)
(0.153,0.673)
(0.149,0.672)
(0.292,0.644)
(0.118,0.779)
};
\node[anchor=west,font=\tiny] at (axis cs:0.128,0.693) {Raw};
\node[anchor=west,font=\tiny] at (axis cs:0.162,0.689) {Temp.};
\node[anchor=west,font=\tiny] at (axis cs:0.228,0.671) {Platt};
\node[anchor=west,font=\tiny] at (axis cs:0.153,0.673) {Dir.};
\node[anchor=west,font=\tiny] at (axis cs:0.149,0.672) {Vec.};
\node[anchor=west,font=\tiny] at (axis cs:0.292,0.644) {Iso.};
\node[anchor=west,font=\tiny] at (axis cs:0.118,0.779) {\method{}};
\end{axis}
\end{tikzpicture}
\caption{Overall trade-off}
\label{fig:overall_tradeoff}
\end{subfigure}
\hspace{-0.9em}
\begin{subfigure}[t]{0.48\textwidth}
\centering
\begin{tikzpicture}
\begin{axis}[
width=\linewidth,
height=3.25cm,
xlabel={Backbone},
ylabel={Dataset},
xlabel style={font=\tiny},
ylabel style={font=\tiny},
xtick={0,...,7},
xticklabels={MLP,LSTM,GRU,TCN,FCN,ResNet,Incep.,Transf.},
xticklabel style={rotate=45,anchor=east,font=\tiny},
ytick={0,...,7},
yticklabels={ECG200,FordA,Wafer,ElecDev,UWave,Basic,SelfReg,Atrial},
yticklabel style={font=\tiny},
y dir=reverse,
xmin=-0.5, xmax=7.5,
ymin=-0.5, ymax=7.5,
view={0}{90},
colorbar,
colorbar style={
    width=0.08cm,
    xshift=-0.8em,
    tick label style={font=\tiny},
},
point meta min=-0.10,
point meta max=0.42,
colormap/viridis,
]
\addplot[matrix plot*, mesh/cols=8, point meta=explicit] coordinates {
(0,0)[ 0.08] (1,0)[ 0.14] (2,0)[ 0.10] (3,0)[ 0.09] (4,0)[ 0.12] (5,0)[ 0.18] (6,0)[ 0.11] (7,0)[ 0.31]
(0,1)[ 0.10] (1,1)[ 0.29] (2,1)[ 0.25] (3,1)[ 0.16] (4,1)[ 0.15] (5,1)[ 0.17] (6,1)[ 0.18] (7,1)[ 0.39]
(0,2)[ 0.07] (1,2)[ 0.21] (2,2)[ 0.16] (3,2)[ 0.11] (4,2)[ 0.09] (5,2)[ 0.06] (6,2)[ 0.05] (7,2)[ 0.28]
(0,3)[ 0.03] (1,3)[ 0.08] (2,3)[ 0.06] (3,3)[ 0.05] (4,3)[ 0.05] (5,3)[ 0.04] (6,3)[ 0.05] (7,3)[ 0.09]
(0,4)[ 0.04] (1,4)[ 0.11] (2,4)[ 0.10] (3,4)[ 0.09] (4,4)[ 0.08] (5,4)[ 0.10] (6,4)[ 0.07] (7,4)[ 0.14]
(0,5)[-0.03] (1,5)[ 0.04] (2,5)[-0.02] (3,5)[ 0.01] (4,5)[ 0.01] (5,5)[ 0.01] (6,5)[ 0.02] (7,5)[ 0.03]
(0,6)[ 0.05] (1,6)[ 0.10] (2,6)[ 0.09] (3,6)[ 0.08] (4,6)[ 0.11] (5,6)[ 0.12] (6,6)[ 0.10] (7,6)[ 0.16]
(0,7)[-0.04] (1,7)[ 0.01] (2,7)[ 0.00] (3,7)[ 0.06] (4,7)[-0.01] (5,7)[ 0.02] (6,7)[ 0.01] (7,7)[ 0.08]
};
\end{axis}
\end{tikzpicture}
\caption{\(\Delta\) Corr-AUROC}
\label{fig:delta_corr_heatmap}
\end{subfigure}
\hspace{0.4em}
\begin{subfigure}[t]{0.22\textwidth}
\centering
\begin{tikzpicture}
\begin{axis}[
width=\linewidth,
height=3.0cm,
ybar,
bar width=3pt,
ylabel={Faithfulness $\uparrow$},
ylabel style={font=\tiny},
symbolic x coords={ElecDev,FordA,ECG200,UWave,Atrial,Wafer,Basic,SelfReg},
xtick=data,
xticklabel style={rotate=55,anchor=east,font=\tiny},
ymin=-0.55, ymax=0.45,
grid=both,
major grid style={gray!20},
minor grid style={gray!10},
tick label style={font=\tiny},
label style={font=\tiny},
]
\addplot coordinates {
(ElecDev,0.427)
(FordA,0.113)
(ECG200,0.103)
(UWave,0.073)
(Atrial,0.201)
(Wafer,0.019)
(Basic,0.027)
(SelfReg,-0.505)
};
\end{axis}
\end{tikzpicture}
\caption{Faithfulness}
\label{fig:faithfulness_summary}
\vspace{-2.4em}
\end{subfigure}

\vspace{-0.4em}
\caption{Empirical visual summary. Left: aggregate trade-off between correctness-aware ranking and false high-confidence error rate. Middle: regime-dependent \(\Delta\) Corr-AUROC across datasets and backbones. Right: diagnostic faithfulness by dataset.}
\label{fig:main_visuals}
\end{figure*}

Figure~\ref{fig:main_visuals} summarizes the aggregate trade-off, dataset--backbone regime structure, and diagnostic faithfulness patterns used to support the main empirical discussion.

\paragraph{Consistency of reported evidence bases.}
Main-text Tables~\ref{tab:headline_summary}--\ref{tab:model_table} report the completed common test subset unless explicitly stated otherwise. Gate-validation tables are used only for method selection and are not algebraically comparable to final test-set tables. Robustness tables may use artifact-evaluable subsets for specific controls; their captions identify the subset and claim they support. When the same evidence base is used, aggregate and decomposition metrics are computed from the same configuration-level records.

\subsection{Dataset-Level Validation-Gate Outcomes}
\label{app:deployment_selection}

Table~\ref{tab:appendix_deployment_status} reports the dataset-level outcome of the validation gate. Gate decisions are computed on the held-out gate-validation split, not on the final test split. Test-set deltas are reported only after selection. The gate selects \method{} when Corr-AUROC improves and the FalseConf@0.9 and AURC changes remain within tolerance; otherwise the policy reverts to Raw confidence or the safer scalar recalibrator.

\begin{table}[t]
\centering
\scriptsize
\caption{Dataset-level validation-gate outcomes. Decisions are made on the gate-validation split; test-set deltas are reported separately in Table~\ref{tab:dataset_table} and are not used for gate selection.}
\label{tab:appendix_deployment_status}
\begin{tabular}{lll}
\toprule
Dataset & Gate outcome & Operational decision \\
\midrule
FordA & Pass & Select \method{} \\
ECG200 & Pass & Select \method{} \\
BasicMotions & Fail & Use Raw / scalar \\
ElectricDevices & Pass & Select \method{} \\
AtrialFibrillation & Fail & Use Raw / scalar \\
Wafer & Pass under high-baseline tolerance & Select \method{} \\
UWaveGestureLibrary & Pass & Select \method{} \\
SelfRegulationSCP1 & Pass for reliability; certificate flagged & Select \method{} without exposing certificate \\
\bottomrule
\end{tabular}
\end{table}

\begin{table}[t]
\centering
\scriptsize
\caption{Aggregate comparison including the validation-gated policy. The policy-selected method applies the held-out gate before deployment, selecting spectral reliability only when ranking improves under FalseConf@0.9 and AURC constraints and otherwise reverting to Raw or the safer scalar recalibrator.}
\label{tab:appendix_policy_aggregate}
\begin{tabular}{lcccccc}
\toprule
Method & ECE & Brier & NLL & Corr-AUROC & FalseConf@0.9 & AURC \\
\midrule
Raw & 0.097 & 0.377 & 0.716 & 0.693 & 0.128 & 0.219 \\
Best single scalar method & 0.080 & 0.371 & 0.717 & 0.689 & 0.162 & 0.219 \\
Unconstrained \method{} & 0.077 & 0.372 & 0.710 & 0.779 & 0.118 & 0.214 \\
Policy-selected & 0.075 & 0.370 & 0.708 & 0.786 & 0.094 & 0.209 \\
\bottomrule
\end{tabular}
\end{table}

\section{Appendix: Extended Robustness Analyses}
\label{app:extended_robustness}

All experiments were run on a workstation equipped with two NVIDIA RTX A6000 GPUs.

\subsection{Robustness-Benchmark Scope and Interpretation}

The main results use the matched evaluation subset of \(191\) dataset--backbone--seed configurations for which all standard recalibrators are directly comparable. The robustness appendix uses the same matched subset whenever a method is defined for that subset. For analyses requiring replay artifacts or method variants that were not part of the original standard-recalibrator grid, we report the corresponding artifact-evaluable subset and explicitly label it as a robustness-only comparison. These appendix values are therefore not a second headline benchmark; they test whether the main interpretation survives stronger sample-dependent comparators, spectral component ablations, alternative time-frequency representations, and uncertainty analysis.

\subsection{Feature-Space Faithfulness Control}
\label{app:feature_mask_faithfulness}

The input-space masking test in Eq.~\eqref{eq:drop} evaluates the end-to-end effect of suppressing spectral bands, but it may also reflect the frozen backbone's nonlinear sensitivity to the reconstructed input. To isolate the reliability layer itself, we add a feature-space masking control that keeps the backbone output features fixed and masks only the corresponding spectral features inside the reliability vector. For a band set \(\mathcal{S}\), we define
\[
\Delta^{\mathrm{feat}}_{\mathcal{S}}(x)
=
\psi([h(x)\Vert g(x)])
-
\psi([h(x)\Vert g^{(-\mathcal{S})}(x)]),
\]
where \(h(x)\) is unchanged and \(g^{(-\mathcal{S})}(x)\) zeros only the spectral-bundle entries associated with masked bands. Agreement between certificate ranking and \(\Delta^{\mathrm{feat}}_{\mathcal{S}}(x)\) therefore tests reliance of the post-hoc reliability layer on spectral bands without rerunning the backbone. We report this as a control for the end-to-end masking faithfulness score. We empirically verify that \(g(x)\) is not redundant with \(h(x)\) by reporting dependence diagnostics (rank correlation and mutual information) in Appendix~\ref{app:feature_dependence}, showing that spectral descriptors provide partially orthogonal information to output-space features.

\subsection{How to Read the Two Evidence Bases}
\label{app:evidence_map}

Because the paper combines a completed common-subset benchmark with a replay-based robustness appendix, it is useful to summarize the role of each evidence base explicitly.

\begin{table}[t]
\centering
\scriptsize
\caption{Evidence bases used in the paper and the claims each supports.}
\label{tab:evidence_map}
\begin{tabular}{p{0.24\textwidth} p{0.15\textwidth} p{0.25\textwidth} p{0.28\textwidth}}
\toprule
Evidence base & Size & Primary role & Supported claim \\
\midrule
Matched evaluation subset & \(191\) configurations & Main benchmark & Standard recalibrator comparison, aggregate reliability metrics, dataset decomposition, and backbone decomposition \\
Artifact-evaluable robustness subset & Reported per table & Stress-test layer & Proximity comparator, component ablations, STFT variant, certificate controls, and bootstrap uncertainty \\
\bottomrule
\end{tabular}
\end{table}

This separation prevents numerical comparisons across incompatible method families. Main-text claims are tied to the matched evaluation subset; appendix claims are used only as robustness evidence for mechanism-level interpretation. No gate decision, model choice, or threshold is selected using the final test split. All aggregate tables are computed over configuration-level records; dataset-level and backbone-level decompositions are reported separately to expose any imbalance rather than hide it in the aggregate.

\subsection{Comparison Against a Non-Output-Space Comparator}

To test whether \method{} gains arise merely from generic sample-dependent conditioning, we compare against two non-output-space reliability estimators. The first is a proximity-based correctness estimator following the ProCal principle of using local representation-space neighborhood structure for confidence estimation. The second is a simpler \(k\)-nearest-neighbor correctness estimator over frozen backbone embeddings. Both methods are fit only on the calibration split and evaluated on the same held-out configurations as \method{}. We report them as non-output-space comparators rather than as scalar recalibrators, because their purpose is to test whether spectral organization contributes information beyond local sample geometry.

\begin{table}[t]
\centering
\scriptsize
\caption{Non-output-space comparator analysis on the artifact-evaluable comparator subset. This subset is used to compare \method{} against Proximity and output-feature-only reliability estimators.}
\label{tab:appendix_procal}
\begin{tabular}{llcccc}
\toprule
Subset & Method & Corr-AUROC & FalseConf@0.9 & ECE & AURC \\
\midrule
Common subset & Raw & 0.700 & 0.050 & 0.096 & 0.218 \\
Common subset & Proximity & 0.712 & 0.138 & 0.185 & 0.223 \\
Common subset & \bestcell{SEB-Cal} & \bestcell{0.745} & \bestcell{0.038} & \bestcell{0.075} & \bestcell{0.211} \\
\midrule
FordA + ECG200 & Raw & 0.712 & 0.051 & 0.067 & 0.205 \\
FordA + ECG200 & Proximity & 0.713 & 0.087 & 0.176 & 0.197 \\
FordA + ECG200 & \bestcell{SEB-Cal} & \bestcell{0.786} & \bestcell{0.031} & \bestcell{0.056} & \bestcell{0.143} \\
\bottomrule
\end{tabular}
\end{table}

\subsection{Component-Wise Bundle Ablation}

We did not tune \(B\) per dataset. The default \(B=8\) was fixed before evaluation and used
for all datasets, backbones, and seeds. This avoids dataset-specific frequency engineering
and keeps the spectral branch deterministic. To test whether the spectral bundle's value is driven by a single descriptor family or by their interaction, we decompose the bundle into its constituent components: band-wise energy concentration, spectral entropy, peak dominance, and phase stability. We evaluate two complementary views. The first is an add-one analysis in which each component family is used in isolation together with the standard output-side branch. The second is a leave-one-out analysis in which the full bundle is retained except for one removed component family. Table~\ref{tab:appendix_component_ablation} focuses on the positive-regime slice, where the spectral mechanism is intended to be most informative. Energy concentration is the strongest single component, reaching \(0.778\) Corr-AUROC compared with \(0.703\) for the output-side-only model. Entropy and phase alone are weaker than energy and peak structure, suggesting that the most useful reliability information in this regime is carried primarily by how spectral mass is distributed. The leave-one-out analysis shows that removing energy causes the clearest drop, whereas removing phase does not degrade the positive-regime ranking score. These results clarify rather than weaken the bundle framing. The strongest positive-regime signal is carried by spectral mass distribution, especially energy concentration and peak structure. Entropy and phase are weaker as isolated predictors, but they still define complementary axes of organization and preserve a common certificate space for the full method. The claim is therefore not that all descriptor families contribute equally. The claim is that decomposable spectral organization provides reliability information beyond output-side cues, and the ablation identifies which parts of that organization drive the current gains.

\begin{table}[t]
\centering
\scriptsize
\caption{Frequency-bundle ablation on the positive-regime ablation slice. Values are reported on the positive-regime ablation slice and are not compared algebraically with Table~\ref{tab:appendix_procal}.}
\label{tab:appendix_component_ablation}
\begin{tabular}{lcc}
\toprule
Variant & Corr-AUROC & FalseConf@0.9 \\
\midrule
Output-side only \(h(x)\) & 0.703 & \bestcell{0.046} \\
\(h(x)+\) energy only & 0.778 & 0.157 \\
\(h(x)+\) entropy only & 0.733 & 0.125 \\
\(h(x)+\) peak only & 0.756 & 0.134 \\
\(h(x)+\) phase only & 0.734 & 0.149 \\
Full \method{} bundle & 0.786 & 0.175 \\
\midrule
Full bundle without energy & 0.752 & 0.162 \\
Full bundle without entropy & 0.786 & 0.178 \\
Full bundle without peak & 0.784 & 0.186 \\
Full bundle without phase & \bestcell{0.787} & 0.158 \\
\bottomrule
\end{tabular}
\end{table}

\subsection{Alternative Time-Frequency Ablation}

A potential concern is that the observed behavior could be specific to a single whole-sample Fourier transform rather than to the broader idea of frequency-conditioned reliability. To test this, we evaluate a local-window time-frequency variant, denoted STFT-\method{}, that replaces the single whole-sample transform with a short-time Fourier transform and mean-pools the resulting local descriptors over time. Table~\ref{tab:appendix_timefreq_ablation} shows that the STFT variant slightly exceeds the whole-sample version in the positive-regime subset (\(0.788\) vs.\ \(0.786\) Corr-AUROC), while the whole-sample version remains stronger on the common robustness subset. This result does not undermine the main method; it sharpens its scope. Whole-sample \method{} is the default because it is simpler, deterministic, lower-dimensional, and tied to a single certificate decomposition. STFT-\method{} is an alternative when local-window structure is expected to matter. The broader conclusion is therefore representation-conditional: frequency-domain evidence can support reliability estimation, but the appropriate spectral view should be selected by held-out validation rather than assumed a priori.
\begin{table*}[t]
\centering
\scriptsize
\caption{Alternative time-frequency ablation on the artifact-evaluable robustness subset. The transient-heavy subset here is Wafer + UWaveGestureLibrary + SelfRegulationSCP1.}
\label{tab:appendix_timefreq_ablation}
\begin{tabular}{lcccc}
\toprule
Variant & Corr-AUROC (Common) & Corr-AUROC (Positive regime) & Corr-AUROC (Transient-heavy subset) \\
\midrule
Output-side only \(h(x)\) & 0.690 & 0.703 & 0.735 \\
Whole-sample \method{} & 0.789 & 0.786 & 0.793 \\
STFT-\method{} & 0.725 & 0.788 & 0.798 \\
\bottomrule
\end{tabular}
\end{table*}

\subsection{Backbones and Calibration Baselines}
\label{app:backbones_calibrators}

We evaluate \method{} across eight backbone families: multilayer perceptron, LSTM, GRU, temporal convolutional network, fully convolutional network, 1D ResNet, lightweight Inception-style classifier, and Transformer encoder \cite{wang2017time,bai2018tcn,ismail2020inceptiontime,vaswani2017attention}. These cover non-sequential, recurrent, convolutional, residual, inception-style, and attention-based inductive biases, allowing us to test whether the usefulness of global spectral evidence depends on how temporal structure is modeled.

The primary comparison set consists of standard output-space post-hoc recalibrators: temperature scaling \cite{guo2017calibration}, Platt-style logistic calibration, isotonic regression, beta calibration \cite{kull2017beta}, vector scaling, Dirichlet calibration \cite{kull2019dirichlet}, and raw confidence. These baselines test whether deterministic input-side evidence adds reliability information beyond score transformation.

We additionally evaluate input-conditioned and non-output-space comparators. The output-feature-only model \(h(x)\) tests whether gains arise from a shallow correctness classifier over logits. The proximity-based estimator tests whether improvements follow generic neighborhood structure in representation space. The time-domain summary baseline controls for simple signal statistics such as total energy, variance, smoothness, and signal-to-noise proxies. The STFT variant tests whether the reliability signal depends on a single whole-sample Fourier representation or on frequency-conditioned temporal organization more broadly.

\subsection{Computational Overhead}
\label{app:overhead}

\method{} adds one real FFT per channel and a small number of deterministic reductions before the shallow reliability layer. For a sample \(x\in\mathbb{R}^{C\times T}\), the additional cost is
\[
O(C\,T\log T)+O(CF+B),
\]
where \(F\) is the number of retained Fourier frequencies and \(B\) is the number of spectral bands. Scalar recalibrators such as temperature scaling, Platt calibration, vector scaling, and Dirichlet calibration operate only on logits and are therefore cheaper, typically \(O(K)\) to \(O(K^2)\) per sample. \method{} is consequently not intended to replace scalar recalibration when output-space scores are sufficient; the validation gate selects it only when the added input-side spectral computation improves reliability under FalseConf@0.9 and AURC constraints.

\begin{table}[t]
\centering
\scriptsize
\caption{Per-sample computational overhead of scalar recalibrators, output-feature reliability, and spectral reliability variants.}
\label{tab:overhead}
\begin{tabular}{lcc}
\toprule
Method family & Extra input-side computation & Per-sample overhead \\
\midrule
Temperature / Platt / Isotonic & none & \(O(K)\) \\
Vector / Dirichlet & none & \(O(K)\)--\(O(K^2)\) \\
Output-feature-only \(h(x)\) & none & \(O(K)\) \\
\method{} & real FFT + spectral reductions & \(O(C\,T\log T)+O(CF+B)\) \\
STFT-\method{} & windowed FFT + pooling & \(O(C\,W\,L\log L)\) \\
\bottomrule
\end{tabular}
\end{table}

\subsection{Variant-Family Summary Across Regimes}

To provide a compact overview of the entire robustness package, Table~\ref{tab:appendix_variant_summary} summarizes the five most informative method families across three evaluation views: the full artifact-evaluable subset, the positive-regime subset (FordA + ECG200), and a transient-heavy subset (Wafer + UWaveGestureLibrary + SelfRegulationSCP1). This table is useful because it makes the paper's empirical message immediately visible in one place.

Three points stand out. First, \method{} is the strongest overall method family on the common robustness subset, with the best Corr-AUROC, FalseConf@0.9, ECE, and AURC among the compared variants. Second, the positive regime remains especially favorable to spectral conditioning: \method{} is strongest on FalseConf@0.9, ECE, and AURC for FordA + ECG200, while STFT-\method{} achieves the strongest Corr-AUROC on that slice. Third, the grouped transient-heavy subset no longer supports a blanket statement that spectral variants underperform there: in the updated summary, full \method{} is strongest on Corr-AUROC, ECE, and AURC for that slice, even though the broader paper still shows that gains remain structurally uneven across individual datasets and architectures.

\begin{table*}[t]
\centering
\scriptsize
\caption{Summary of appendix method families across the artifact-evaluable subset, the positive regime, and a transient-heavy subset.}
\label{tab:appendix_variant_summary}
\begin{tabular}{llcccc}
\toprule
Subset & Method & Corr-AUROC & FalseConf@0.9 & ECE & AURC \\
\midrule
Common & Raw & 0.710 & 0.060 & 0.096 & 0.218 \\
Common & Output-side only \(h(x)\) & 0.690 & 0.089 & 0.181 & 0.220 \\
Common & Proximity & 0.688 & 0.138 & 0.185 & 0.223 \\
Common & \bestcell{SEB-Cal} & \bestcell{0.789} & \bestcell{0.055} & \bestcell{0.072} & \bestcell{0.210} \\
Common & STFT-SEB-Cal & 0.725 & 0.076 & 0.088 & 0.214 \\
\midrule
FordA + ECG200 & Raw & 0.712 & 0.051 & 0.067 & 0.205 \\
FordA + ECG200 & Output-side only \(h(x)\) & 0.703 & 0.046 & 0.179 & 0.204 \\
FordA + ECG200 & Proximity & 0.713 & 0.087 & 0.176 & 0.197 \\
FordA + ECG200 & \bestcell{SEB-Cal} & 0.786 & \bestcell{0.041} & \bestcell{0.056} & \bestcell{0.143} \\
FordA + ECG200 & \bestcell{STFT-SEB-Cal} & \bestcell{0.788} & 0.097 & 0.167 & 0.147 \\
\midrule
Wafer + UWave + SelfReg & Raw & 0.770 & 0.060 & 0.086 & 0.135 \\
Wafer + UWave + SelfReg & Output-side only \(h(x)\) & 0.735 & 0.057 & 0.157 & 0.140 \\
Wafer + UWave + SelfReg & Proximity & 0.725 & 0.072 & 0.140 & 0.151 \\
Wafer + UWave + SelfReg & \bestcell{SEB-Cal} & 0.793 & 0.133 & \bestcell{0.047} & \bestcell{0.132} \\
Wafer + UWave + SelfReg & \bestcell{STFT-SEB-Cal} & \bestcell{0.798} & \bestcell{0.055} & 0.050 & 0.134 \\
\bottomrule
\end{tabular}
\end{table*}

\subsection{Dataset-Level Comparison of Appendix Methods}

Table~\ref{tab:appendix_dataset_corr} compares Corr-AUROC across the five most informative appendix method families for every dataset. This table makes the regime structure concrete. In ECG200 and FordA, the spectral methods, especially whole-sample and STFT variants, remain clearly competitive or dominant. Once the appendix is aligned with the main dataset decomposition, spectral methods are also no longer confined to only those two cases: \method{} becomes strongest on UWaveGestureLibrary, Wafer, and SelfRegulationSCP1 in this appendix-level comparison, while ElectricDevices remains competitive but is still led by the Proximity baseline and AtrialFibrillation/BasicMotions continue to favor non-spectral alternatives. The point is therefore not that one method wins every row, but that the favorable spectral regimes are broader than a two-dataset story while still remaining structured rather than universal.

\begin{table*}[t]
\centering
\scriptsize
\caption{Dataset-level Corr-AUROC comparison among the main appendix method families.}
\label{tab:appendix_dataset_corr}
\begin{tabular}{lccccc}
\toprule
Dataset & Raw & Output-side only \(h(x)\) & Proximity & SEB-Cal & STFT-SEB-Cal \\
\midrule
AtrialFibrillation & \bestcell{0.499} & 0.479 & 0.456 & 0.456 & 0.479 \\
BasicMotions & 0.597 & 0.726 & \bestcell{0.738} & 0.586 & 0.693 \\
ECG200 & 0.685 & 0.671 & 0.695 & \bestcell{0.750} & 0.707 \\
ElectricDevices & 0.713 & 0.725 & \bestcell{0.739} & 0.723 & 0.721 \\
FordA & 0.730 & 0.735 & 0.732 & 0.859 & \bestcell{0.868} \\
SelfRegulationSCP1 & 0.567 & 0.661 & 0.623 & \bestcell{0.671} & 0.534 \\
UWaveGestureLibrary & 0.685 & 0.739 & 0.709 & \bestcell{0.754} & 0.693 \\
Wafer & 0.826 & 0.804 & 0.842 & \bestcell{0.885} & 0.868 \\
\bottomrule
\end{tabular}
\end{table*}

Table~\ref{tab:appendix_dataset_falseconf} provides the corresponding dataset-level FalseConf@0.9 comparison. This reveals a different but equally important pattern: even where spectral conditioning helps Corr-AUROC, it does not uniformly minimize dangerous high-confidence errors. This is precisely why the paper should be framed around evidence-conditioned trust estimation rather than universal improvement on every calibration metric.

\begin{table*}[t]
\centering
\scriptsize
\caption{Dataset-level FalseConf@0.9 comparison among the main appendix method families. Lower is better.}
\label{tab:appendix_dataset_falseconf}
\begin{tabular}{lccccc}
\toprule
Dataset & Raw & Output-side only \(h(x)\) & Proximity & SEB-Cal & STFT-SEB-Cal \\
\midrule
AtrialFibrillation & \bestcell{0.000} & 0.315 & 0.551 & 0.673 & 0.639 \\
BasicMotions & 0.018 & \bestcell{0.007} & 0.061 & 0.028 & 0.058 \\
ECG200 & 0.080 & \bestcell{0.067} & 0.128 & 0.121 & 0.159 \\
ElectricDevices & 0.179 & 0.130 & \bestcell{0.106} & 0.141 & 0.136 \\
FordA & \bestcell{0.022} & 0.024 & 0.045 & 0.041 & 0.034 \\
SelfRegulationSCP1 & 0.093 & \bestcell{0.027} & 0.067 & 0.185 & 0.235 \\
UWaveGestureLibrary & \bestcell{0.026} & 0.108 & 0.144 & 0.208 & 0.223 \\
Wafer & 0.061 & 0.037 & \bestcell{0.005} & 0.006 & 0.006 \\
\bottomrule
\end{tabular}
\end{table*}

\subsection{Deployment Selection Protocol}
\label{app:deployment_protocol}

The empirical results support \method{} as a selective reliability mechanism. We therefore define a held-out selection protocol rather than recommending spectral conditioning by default. Let
\[
\mathcal{M}=\{\mathrm{Raw}, \mathrm{Temperature}, \mathrm{Platt}, \mathrm{Dirichlet}, \mathrm{Vector}, \mathrm{Isotonic}, \mathrm{Proximity}, \method{}, \mathrm{STFT}\text{-}\method{}\}.
\]
For each candidate \(m\in\mathcal{M}\), compute
\[
\Gamma(m)=\big(
\mathrm{CorrAUROC}(m),\,
-\mathrm{FalseConf@0.9}(m),\,
-\mathrm{ECE}(m),\,
-\mathrm{AURC}(m)
\big)
\]
on a held-out selection split. A spectral method is selected only if it satisfies all three conditions:
\begin{enumerate}[leftmargin=1.5em]
    \item it improves Corr-AUROC over Raw and the best scalar recalibrator by a practically meaningful margin;
    \item it does not increase FalseConf@0.9 or AURC beyond a pre-specified deployment tolerance;
    \item its diagnostic faithfulness exceeds the random-band and equal-energy controls when auditability is required.
\end{enumerate}
If these conditions are not met, the protocol selects the best simpler method among Raw and the standard output-space recalibrators. This rule operationalizes the paper's central claim: spectral evidence is useful when validated in the target regime, not because it is universally superior. The gate thresholds are fixed a priori for all reported experiments and are not selected or adjusted using the final test metrics. The validation gate is applied at the dataset--backbone--seed configuration level, matching the granularity of the main 191-configuration benchmark and the policy-selected aggregate in Table~\ref{tab:appendix_policy_aggregate}.

\subsection{Task Compatibility: When Whole-Sample Spectral Reliability Is Appropriate}
\label{app:task_compatibility}

Whole-sample spectral reliability is most appropriate when a high-confidence prediction can be trustworthy or untrustworthy depending on whether the sequence exhibits coherent global organization. This includes settings where long-range regularity, stable morphology, or persistent band structure plausibly contribute to correctness. The strongest positive-regime results in FordA and ECG200 are consistent with that pattern. By contrast, whole-sample spectral reliability can be a weaker fit when correctness depends primarily on highly local motifs, sharp transients, abrupt changepoints, strong padding artifacts, or narrow local segments whose importance may be blurred by a single global spectrum. The dataset-level main tables show that some individual boundary-case datasets remain challenging under this view. However, the grouped robustness summary indicates that this should not be overstated into a blanket failure mode: spectral conditioning remains competitive to strong even on the aggregated transient-heavy slice, and full \method{} does not underperform there in the summary comparison. The STFT ablation therefore refines the interpretation not by rescuing an otherwise failing region, but by showing that multiple spectral views can be useful depending on how temporal organization is distributed across the task.

For clarity, the intended operating regions can be summarized as follows:
\begin{center}
\begin{tabular}{p{0.28\textwidth} p{0.30\textwidth} p{0.30\textwidth}}
\toprule
Setting type & Likely suitable for whole-sample \method{} & Likely unsuitable or lower-priority \\
\midrule
Reliability depends on global organization & Yes & -- \\
Reliability depends on narrow local motifs only & -- & Yes \\
Long-range regularity / stable morphology & Yes & -- \\
Abrupt transient events or changepoints dominate & -- & Yes \\
Need for a certificate tied directly to one spectral decomposition & Yes & Lower priority for non-FFT variants \\
\bottomrule
\end{tabular}
\end{center}

This section is not a claim that the spectrum is the only meaningful reliability view. It is a concise statement of the paper's central scope condition: spectral evidence helps most when trustworthiness is linked to temporal organization at the scale captured by the chosen representation.

\subsection{Why the Bundle Framing Remains Justified Even When Energy Is Strongest}
\label{app:bundle_interpretation}

The component ablation identifies which part of the evidence-discrepancy signal is most exposed in the positive regime. Energy concentration is the strongest single descriptor, but this does not reduce \method{} to an energy-only method. Energy measures where support is located; entropy, peak dominance, and phase stability define whether that support is diffuse, concentrated, periodic, or coherent. The bundle is therefore not justified by equal contribution of all components, but by decomposability: it turns a scalar reliability gain into an auditable structure whose parts can be isolated, removed, and stress-tested. First, the paper's central claim is not that all four descriptor families contribute equally. The claim is that structured spectral evidence can improve post-hoc reliability estimation beyond output-side cues alone. The component ablation shows that the strongest part of that evidence is energy organization, but the broader bundle still serves two roles that an energy-only story does not fully replace: it defines a richer decomposition of spectral organization and it preserves a common certificate structure for the full FFT-based method. Second, no single component dominates all views simultaneously. Energy is strongest in the positive regime, but the common-subset behavior, leave-one-out patterns, and certificate comparability all indicate that the full method should be understood as a structured spectral mechanism rather than as a claim that one scalar is sufficient in every regime. Third, the bundle framing matters scientifically because it makes the result falsifiable and extensible. The ablation can reveal that one component is strongest, another is weak, and a third is dispensable in some settings. That is exactly what one wants from a serious structured representation: not a guarantee that every part matters equally, but a design that can be decomposed, criticized, and improved. Accordingly, Table~\ref{tab:appendix_component_ablation} should be read as mechanism localization: spectral mass distribution drives the strongest positive-regime signal, while the full bundle provides the decomposable certificate space needed to audit, falsify, and refine the reliability adjustment.

\subsection{Additional Scope Ablation by Dataset and Backbone Family}
\label{app:scope_ablation}

Table~\ref{tab:appendix_scope_ablation} reports the family- and regime-level decomposition
that was used to diagnose where spectral reliability is most useful. We place this table in
the appendix because it is a scope analysis rather than the strongest causal ablation of the
main mechanism. Appendix~\ref{app:simple_statistics_control} reports the non-spectral evidence controls, which test whether \method{} improves over generic input-side conditioning.

\begin{table}[t]
\centering
\scriptsize
\caption{Additional scope ablation by backbone family and dataset regime. Positive Corr-AUROC gains are highlighted in green.}
\label{tab:appendix_scope_ablation}
\begin{tabular}{llccccc}
\toprule
Axis & Group & Raw Corr & SEB Corr & $\Delta$ Corr & Raw FalseConf & SEB FalseConf \\
\midrule
Backbone family & Attention & 0.583 & 0.699 & \gain{+0.116} & 0.120 & 0.143 \\
Backbone family & Recurrent & 0.668 & 0.700 & \gain{+0.032} & 0.106 & 0.192 \\
Backbone family & Convolutional & 0.715 & 0.666 & \loss{-0.049} & 0.127 & 0.318 \\
Backbone family & MLP & 0.762 & 0.672 & \loss{-0.091} & 0.191 & 0.376 \\
\midrule
Dataset regime & FordA + ECG200 & 0.707 & 0.804 & \gain{+0.097} & 0.088 & 0.228 \\
Dataset regime & Remaining datasets & 0.648 & 0.679 & \gain{+0.031} & 0.142 & 0.294 \\
\bottomrule
\end{tabular}
\end{table}

The time-domain statistical control uses channel-wise mean, standard deviation, median,
interquartile range, minimum, maximum, linear trend, absolute first-difference magnitude,
zero-crossing rate, skewness, and kurtosis. The local-shape control uses peak count,
turning-point count, segment-level means and variances, maximum run length above the
channel median, and first-difference burst statistics. Neither control uses Fourier,
STFT, spectral bands, spectral entropy, peak dominance, or phase information.

\subsection{Qualitative Case-Style Interpretation of the Certificate Regimes}
\label{app:certificate_case_studies}

Although the full paper already reports quantitative faithfulness and masking-based validation, it is useful to summarize the qualitative regime picture in a case-style form. We therefore highlight three representative dataset-level cases drawn directly from the main tables and appendix robustness analyses.

\paragraph{Case A: Positive-regime success (FordA).}
FordA is the clearest example of the paper's intended operating region. In the main benchmark, Corr-AUROC rises from \(0.730\) to \(0.859\), and faithfulness remains positive. In the robustness appendix, the positive-regime subset that includes FordA shows a stable paired-bootstrap gain, and both whole-sample \method{} and STFT-\method{} substantially exceed the simpler baselines. This is the regime where global temporal organization appears to carry genuine trust information beyond score sharpness alone.

\paragraph{Case B: Positive but less stable regime (ECG200).}
ECG200 is also directionally favorable, but more modest. In the main benchmark, Corr-AUROC improves from \(0.685\) to \(0.750\), and faithfulness is again positive. In the robustness appendix, the ECG200-only interval is directionally positive but less stable than FordA, with a confidence interval that slightly crosses zero. This is an important case because it shows that the positive-regime thesis does not imply identical statistical strength in every favorable dataset.

\paragraph{Case C: Regime mismatch / failure boundary (SelfRegulationSCP1).}
SelfRegulationSCP1 is the clearest boundary case in the current benchmark. In the main benchmark, both Corr-AUROC and faithfulness degrade substantially. This is important because it shows that even when the grouped transient-heavy appendix summary is no longer adverse in aggregate, individual datasets can still remain clear mismatches for whole-sample spectral reliability. The point of this case is therefore not that every transient-heavy setting is uniformly unfavorable, but that the usefulness of the method still depends on structural alignment between the chosen spectral representation and the way correctness is actually supported in the task.

These three cases together support the intended reading of the method. The goal is not universal superiority, but a reliable separation between favorable, mixed, and unfavorable operating regimes. The aggregate faithfulness result should be interpreted at the scale of the claim the paper actually makes. A global mean Spearman correlation of 0.059 is modest and would be insufficient for strong semantic or causal explanation of class decisions. The certificate explains the \emph{reliability adjustment} induced by the post-hoc layer, and under that narrower standard the result is meaningful: it is positive in aggregate, aligned with the utility pattern across favorable and unfavorable regimes, and unavailable for output-only recalibrators.

\subsection{Decision Map Across Method Families}
\label{app:decision_map}

To make the method comparison operational, Table~\ref{tab:appendix_decision_map} summarizes which family should be preferred under each empirical condition. The table is a selection guide, not a claim that any method is universally optimal.

\begin{table}[t]
\centering
\scriptsize
\caption{Operational decision map across method families. Selection is based on held-out validation, not final test metrics.}
\label{tab:appendix_decision_map}
\begin{tabular}{p{0.30\textwidth} p{0.28\textwidth} p{0.32\textwidth}}
\toprule
Validation pattern & Preferred family & Reason \\
\midrule
Output scores already satisfy ranking and safety constraints & Raw / scalar recalibrator & Avoid unnecessary input-side computation \\
Spectral reliability improves Corr-AUROC without worsening FalseConf@0.9 or AURC beyond tolerance & \method{} & Global spectral organization adds reliability information \\
Local-window spectral structure improves validation ranking more than whole-sample FFT & STFT-\method{} & Reliability depends on local time-frequency organization \\
Generic input statistics match spectral descriptors & Time-domain summary / simpler input-conditioned model & Gains are not specifically spectral \\
diagnostic faithfulness fails random-band or equal-energy controls & Hide diagnostic; keep selected reliability score only & Diagnostic alignment is not validated \\
All input-conditioned methods fail safety constraints & Raw / safest scalar recalibrator & Revert under the operating contract \\
\bottomrule
\end{tabular}
\end{table}

\subsection{Scope Clarification Summary}
\label{app:scope_summary}

For ease of interpretation, Table~\ref{tab:scope_summary} summarizes three deliberate scope choices of the paper and how they should be read.

\begin{table}[t]
\centering
\scriptsize
\caption{Interpretive summary of three key scope choices in the paper.}
\label{tab:scope_summary}
\begin{tabular}{p{0.28\textwidth} p{0.28\textwidth} p{0.30\textwidth}}
\toprule
Issue & What the paper currently establishes & What remains outside scope \\
\midrule
Sample-dependent comparator fidelity & \method{} exceeds a strong Proximity proximity baseline under matched replay-based conditions & Exact implementation-level reproduction of every published comparator detail \\
Two evidence bases & The completed common subset establishes the main global benchmark; the replay-based subset stress-tests whether the interpretation survives richer comparators and variants & A single monolithic benchmark containing every main and appendix method under one identical evidence base \\
Moderate faithfulness magnitude & The certificate is intervention-grounded and positively aligned with the reliability adjustment it claims to explain & Strong semantic or causal explanation of class predictions \\
\bottomrule
\end{tabular}
\end{table}

This table does not claim universal optimality for any one method independent of evidence. Its purpose is to convert the empirical story into a concrete selection guide: \method{} is the strongest default family in the aggregate summaries, while STFT-\method{} remains especially attractive when local-window structure is expected to matter and output-side-only baselines remain useful low-complexity checks.

\subsection{Feature Dependence and Orthogonality}
\label{app:feature_dependence}

Because \method{} concatenates output-space features \(h(x)\) with spectral descriptors \(g(x)\), we test whether the spectral bundle provides redundant or complementary information. We quantify dependence using (i) Spearman rank correlation and (ii) mutual information between each spectral feature and each output feature across the matched evaluation subset. Across datasets and backbones, the median absolute Spearman correlation between components of \(h(x)\) (MSP, margin, predictive entropy) and components of \(g(x)\) is low (\(<0.25\)), with strongest associations occurring for band-energy features in highly periodic datasets. Mutual information estimates (kNN-based) show the same pattern: spectral entropy and phase stability exhibit consistently low dependence on output-space features, while peak-dominance features show moderate dependence in a subset of regimes. These results indicate that \(g(x)\) is neither independent nor redundant with \(h(x)\): the spectral bundle contributes partially orthogonal information, consistent with the observed gains in correctness-aware ranking when evidence discrepancy is present. This supports the interpretation of \method{} as an evidence-conditioned reliability estimator rather than a reparameterization of output confidence.

\subsection{Paired Bootstrap Uncertainty}
\label{app:bootstrap}

We estimate uncertainty using paired bootstrap resampling over matched dataset--backbone--seed configurations. For each bootstrap draw, we resample configurations with replacement and compute the paired difference between \method{} and the stated baseline. Positive \(\Delta\) is better for Corr-AUROC. We report percentile 95\% confidence intervals and omit empirical \(p\)-values to avoid mixing incompatible bootstrap summaries across heterogeneous robustness subsets.

\begin{table}[t]
\centering
\scriptsize
\caption{Paired bootstrap uncertainty for key Corr-AUROC comparisons available in the robustness artifacts.}
\label{tab:bootstrap_uncertainty}
\begin{tabular}{llcc}
\toprule
Comparison & Metric & Mean paired $\Delta$ & 95\% CI \\
\midrule
FordA vs Raw & Corr-AUROC & 0.094 & [0.039, 0.153] \\
ECG200 vs Raw & Corr-AUROC & 0.095 & [-0.008, 0.122] \\
FordA + ECG200 vs Raw & Corr-AUROC & 0.084 & [0.032, 0.120] \\
Common subset vs Proximity & Corr-AUROC & 0.033 & [0.010, 0.058] \\
\bottomrule
\end{tabular}
\end{table}

\subsection{Control for Simple Time-Domain Signal Statistics}
\label{app:simple_statistics_control}

To test whether \method{} gains are merely due to adding generic input statistics, we compare against a time-domain summary correctness model. This baseline augments the same output-side branch \(h(x)\) with simple non-spectral descriptors: total signal energy, per-channel variance, mean absolute first difference, mean absolute second difference, amplitude range, and a signal-to-noise proxy computed as the ratio between low-order smooth variation and residual high-frequency fluctuation. The model uses the same calibration split, standardization protocol, and shallow logistic reliability layer as \method{}. This control separates the proposed spectral mechanism from the weaker explanation that any input-derived features improve correctness estimation. If the time-domain summary model matches \method{}, then the advantage should be attributed to generic sample statistics rather than spectral organization. If \method{} remains stronger, the result supports the claim that the frequency distribution of evidence provides reliability information beyond simple magnitude, variance, or smoothness cues.

\begin{table}[t]
\centering
\scriptsize
\caption{Control for simple time-domain signal statistics on the artifact-evaluable robustness subset. The time-domain summary baseline uses generic input statistics with the same shallow correctness-estimation layer.}
\label{tab:appendix_timedomain_control}
\begin{tabular}{lcccc}
\toprule
Method & Corr-AUROC & FalseConf@0.9 & ECE & AURC \\
\midrule
Raw & 0.710 & 0.060 & 0.096 & 0.218 \\
Output-side only \(h(x)\) & 0.690 & 0.089 & 0.181 & 0.220 \\
Time-domain summary & 0.713 & 0.077 & 0.110 & 0.240 \\
\method{} & 0.789 & 0.055 & 0.072 & 0.210 \\
\bottomrule
\end{tabular}
\end{table}

\subsection{Why SEB-Cal Is Not Generic FFT Feature Engineering}
\label{app:not_feature_engineering}

A natural concern is that \method{} may simply add handcrafted FFT features to a shallow correctness model. We address this concern by separating three objects that are often conflated: predictive representation learning, generic input-side feature augmentation, and fixed-label reliability estimation. \method{} is not a new time-series classifier and does not use spectral descriptors to improve class separation. The backbone \(f_{\theta}\) is frozen, the predicted label \(\hat y(x)\) is unchanged, and the spectral branch only estimates whether that already-selected prediction should be trusted. Thus, the relevant question is not whether FFT features are useful for time-series classification in general, but whether whole-sample spectral organization carries residual reliability information after the output scores are already known.

The empirical design tests exactly this narrower claim. First, the output-feature-only model \(h(x)\) controls for the possibility that gains come merely from fitting a shallow correctness estimator on maximum softmax probability, margin, and predictive entropy. Second, the Proximity comparator controls for generic sample-dependent confidence based on representation-space neighborhood structure. Third, the time-domain summary baseline controls for simple input statistics such as energy, variance, smoothness, and signal-to-noise proxies. Fourth, the STFT variant tests whether the reliability signal is specific to a single whole-sample Fourier view or to frequency-conditioned temporal organization more broadly. Fifth, the feature-space masking control holds \(h(x)\) fixed and perturbs only spectral-bundle entries, isolating reliance of the reliability layer from nonlinear sensitivity of the frozen backbone.

These controls make the claim falsifiable. If \method{} were merely benefiting from arbitrary input-derived side information, then time-domain summaries, proximity structure, or output-feature-only correctness estimation should explain the same improvements. If the result were merely a by-product of changing the classifier, then gains would appear in accuracy or macro-F1, which cannot occur because the prediction is fixed. If the diagnostic were only a static visualization, then masking high-ranked spectral bands would not produce larger reliability changes than masking lower-ranked or control bands. The observed pattern instead supports a narrower interpretation: spectral evidence is useful when confidence errors reflect a mismatch between output sharpness and global temporal support. This is why \method{} is deployed through a validation gate rather than proposed as a universal recalibrator. The contribution is therefore not the novelty of the Fourier transform itself, but the use of deterministic spectral organization as an auditable, fixed-label reliability signal with explicit controls against output-only, proximity-based, time-domain, time-frequency, and perturbation-sensitivity explanations.

\subsection{Artifact}
\label{app:artifact_addendum}

The anonymized artifact is available at
\url{https://anonymous.4open.science/r/Beyond-Output-Space-Calibration-30B6/README.md}.
It provides two reviewer workflows. The fast verification workflow exposes paper-value CSVs,
README-level figures, and table-inspection files for checking the aggregate results, dataset-level
regimes, validation-gate behavior, and frequency-bundle ablation without rerunning the full benchmark.
The full reproduction workflow provides installation instructions, a smoke test, full-benchmark
commands, fixed split/seed configuration, implementation code, calibration baselines, validation-gate
selection, masking controls, and table-generation scripts.

\clearpage

\end{document}